\def\year{2022}\relax
\documentclass[letterpaper]{article} 
\usepackage{aaai22}  
\usepackage{times}  
\usepackage{helvet}  
\usepackage{courier}  
\usepackage[hyphens]{url}  
\usepackage{graphicx} 
\usepackage{natbib}  
\usepackage{caption} 
\DeclareCaptionStyle{ruled}{labelfont=normalfont,labelsep=colon,strut=off} 
\usepackage{algorithm}
\usepackage{algorithmic}
\usepackage{subfigure}
\usepackage{amsmath}
\usepackage{paralist}
\usepackage{paralist}
\usepackage{color}
\usepackage{multirow}

\newcommand{\ie}{\textit{i}.\textit{e}.}
\newcommand{\eg}{\textit{e}.\textit{g}.}

\usepackage{newfloat}
\usepackage{listings}
\floatstyle{ruled}
\newfloat{listing}{tb}{lst}{}
\floatname{listing}{Listing}
\title{Multi-View Graph Representation for Programming \\ Language Processing: 
An Investigation into Algorithm Detection}
\author{
    Ting Long\equalcontrib \textsuperscript{\rm 1},
    Yutong Xie\equalcontrib \textsuperscript{\rm 2},
    Xianyu Chen\textsuperscript{\rm 1},
    Weinan Zhang\textsuperscript{\rm 1}$^\textrm{\dag}$,
    Qinxiang Cao\textsuperscript{\rm 1}, 
    Yong Yu\textsuperscript{\rm 1}\thanks{Corresponding author.}
}
\affiliations{
    \textsuperscript{\rm 1}Department of Computer Science and Engineering, Shanghai Jiao Tong University, China \\
    \textsuperscript{\rm 2}School of Information, University of Michigan, Ann Arbor, MI, USA \\
    
    \{longting,xianyujun,wnzhang,caoqinxiang\}@sjtu.edu.cn, yutxie@umich.edu,  yyu@apex.sjtu.edu.cn

}

\usepackage{bibentry}

\begin{document}

\maketitle

\begin{abstract}
Program representation, which aims at converting program source code into vectors with automatically extracted features, is a fundamental problem in programming language processing (PLP). 
Recent work tries to represent programs with neural networks based on source code structures. 
However, such methods often focus on the syntax and consider only one single perspective of programs, limiting the representation power of models. 
This paper proposes a multi-view graph (MVG) program representation method. MVG pays more attention to code semantics and simultaneously includes both data flow and control flow as multiple views. These views are then combined and processed by a graph neural network (GNN) to obtain a comprehensive program representation that covers various aspects. 
We thoroughly evaluate our proposed MVG approach in the context of algorithm detection, an important and challenging subfield of PLP. Specifically, we use a public dataset \texttt{POJ-104} and also construct a new challenging dataset \texttt{ALG-109} to test our method.
In experiments, MVG outperforms previous methods significantly, demonstrating our model's strong capability of representing source code. 
\end{abstract}

\section{Introduction}
With the advent of \textit{big code}  \cite{allamanis2018survey}, programming language processing (PLP) gains plenty of attention in recent years. PLP aims at assisting computers automatically understanding and analyzing source code, which benefits downstream tasks in software engineering like code retrieval \cite{lv2015codehow,nie2016query}, code annotation \cite{yao2019coacor}, bug predicting and fixing \cite{xia2018measuring,wang2018dynamic}, program translation \cite{chen2018tree,gu2017deepam}. 
To take advantage of deep learning, the program representation problem, \ie, how to convert source code into representational vectors, becomes a critical issue in PLP. 

A great deal of literature devotes its efforts to the problem of program representation. 
Among these works, a majority of them represent source code only based on syntactic information like abstract syntax trees (ASTs) \cite{mou2016convolutional,alon2019code2vec} while ignoring the semantics of programs.
Thus, some researchers propose to include semantic information by adding semantic edges onto ASTs \citep{allamanis2018learning,zhou2019devign}. 
However, the program representation still highly depends on the syntax, and the semantics are relatively underweighted. 
Moreover, in previous methods, information from different aspects like syntax, data flow, and control flow are often mixed up into one single view, making the information hard to be disentangled. 

Therefore in this paper, to address the problem mentioned above, we propose to use a multi-view graph (MVG) representation for source code. 
To obtain a more comprehensive understanding of programs, we consider multiple graph views from various aspects and levels. In particular, we emphasize more on semantics and include the following views in MVG: the data-flow graph (DFG), control-flow graph (CFG), read-write graph (RWG), and a combined graph (CG). Among these views, DFG and CFG are widely used in compiling and traditional program analysis. We construct RWG based on DFG and CFG to capture the relationship between operations and operands. We further include CG, a  combination of the former-mentioned graphs, to have an integral representation of the program. 
We then apply a gated graph neural network (GGNN) \cite{li2016gated} to automatically extract information from the four graph views. 

We validate our proposed MVG method in the context of algorithm detection, which is a fundamental subfield of PLP and aims at identifying the algorithms and data structures that appear in the source code.
The first reason for which we choose this subfield is because of its wide application range: 
the detection results can be used as intermediate information for further program analysis; we might also apply algorithm detection in areas like programming education, \eg, determining which algorithms are mastered by the students.  
In addition to the wide range of applications, algorithm detection is also very challenging and can serve as a benchmark for PLP program representation.
This is because:
\begin{inparaenum}[(1)]
    \item A piece of code can contain multiple different algorithms and data structures;
    \item One algorithm or data structure can have multiple possible implementations (\eg, \texttt{Dynamic Programming}, \texttt{Segment Tree});
    \item Different algorithms can have very similar implementations (\eg, \texttt{Dijkstra's Algorithm} and \texttt{Prim's Algorithm}).
\end{inparaenum}
Under this algorithm detection task, we use two datasets to test our MVG model. The first one is a public dataset \texttt{POJ-104} \citep{mou2016convolutional}. We also create a new dataset \texttt{ALG-109}, which is more challenging than the former one. 
On both two datasets, our MVG model outperforms previous methods significantly, demonstrating the outstanding representation power of MVG. 

In summary, our contributions are as follows: 
\begin{itemize}
    \item We propose the MVG method, which can understand source code from various aspects and levels. Specifically, MVG includes four views in total: the data-flow graph (DFG), control-flow graph (CFG), read-write graph (RWG), and a combined graph (CG);
    \item We create an algorithm classification dataset \texttt{ALG-109} to serve as a program representation benchmark;
    \item We validate MVG on the challenging algorithm detection task with a public dataset \texttt{POJ-104} and our constructed dataset \texttt{ALG-109}. In experiments, MVG achieves state-of-the-art performance, illustrating the effectiveness of our approach. 
\end{itemize}

\section{Related Work}
Previous methods on program representation can be divided into four categories:
\textit{data-based}, \textit{sequence-based}, \textit{tree-based}, and \textit{graph-based}.  

\textit{Data-based methods} assume programs are functions that map inputs to outputs. Therefore, such methods use the input and output data to represent the program. 
\citet{piech2015learning} embed inputs and outputs of programs into a vector space and use the embedded vectors to obtain program representations; \citet{wang2019learning} collects all the data during program execution and feeds the data to a long short-term memory (LSTM) unit  \cite{hochreiter1997long} to obtain program representations. 
Though seemingly intuitive, data-based methods are often limited by the availability of the input or output data, and it might take forever to enumerate all possible inputs.

\textit{Sequence-based methods} assume that programming language is similar to natural language, and adjacent units in code (\eg, tokens, instructions, or command lines) will have a strong correlation. Hence, these methods apply models in natural language processing (NLP) to source code. 
For examples, \citet{harer2018automated},  \citet{ben2018neural}, and  \citet{zuo2019neural} apply the word2vec model \cite{le2014distributed} to learn the embeddings of program tokens. 
\citet{feng2020codebert},\citet{wang2020fret} and \citet{ciniselli2021empirical} use a pre-trained BERT model to encode programs. 
Such sequence-based methods are easy to use and can benefit largely from the NLP community. 
However, since source code is highly structured, simple sequential modeling can result in a great deal of information loss. 

\textit{Tree-based methods} are mostly based on the abstract syntax tree (AST), which is often used in compiling. In contrast to the sequential modeling of programming language, AST contains more structural information of source code. In the previous work,
\citet{mou2016convolutional} parse programs into ASTs and then obtain  program representations by applying a tree-based convolutional neural network on the ASTs; 
\citet{alon2019code2vec} obtain program representations by aggregating paths on the AST.
Tree-based representations usually contain more structural information than sequences, but the program semantics might be relatively ignored compared with the syntactic information.

\textit{Graph-based methods} parse programs into graphs. Most approaches from this category construct program graphs by adding edges onto ASTs. 
For instance,  \citet{allamanis2018learning} introduce edges like \texttt{LastRead} and \texttt{LastWrite} into AST. Then the program representations are obtained with a gated graph neural network (GGNN) \cite{li2016gated}.
\citet{zhou2019devign} extend the edges types in the work of \citet{allamanis2018learning} and further improve the performance.
Although graph-based methods can have better performance than previously mentioned categories \cite{allamanis2018learning}, we notice that information from different perspectives usually crowds in one single view, \ie, most methods use one single graph to include various information in the source code, which can limit the representation power of the model. 
Moreover, such approaches tend to build their graphs based on the AST and give much attention to the syntactic information, suppressing the semantics of programs. 
By contrast, in this paper, we propose the MVG method, which considers multiple program views simultaneously. We extract features from data flows, control flows, and read-write flows, focusing more on the semantic elements. 

\section{Methodology}

\begin{figure*}[t]
    \centering
    \subfigure[The MVG pipeline.]{
        \centering 
        \includegraphics[width=0.58\textwidth]{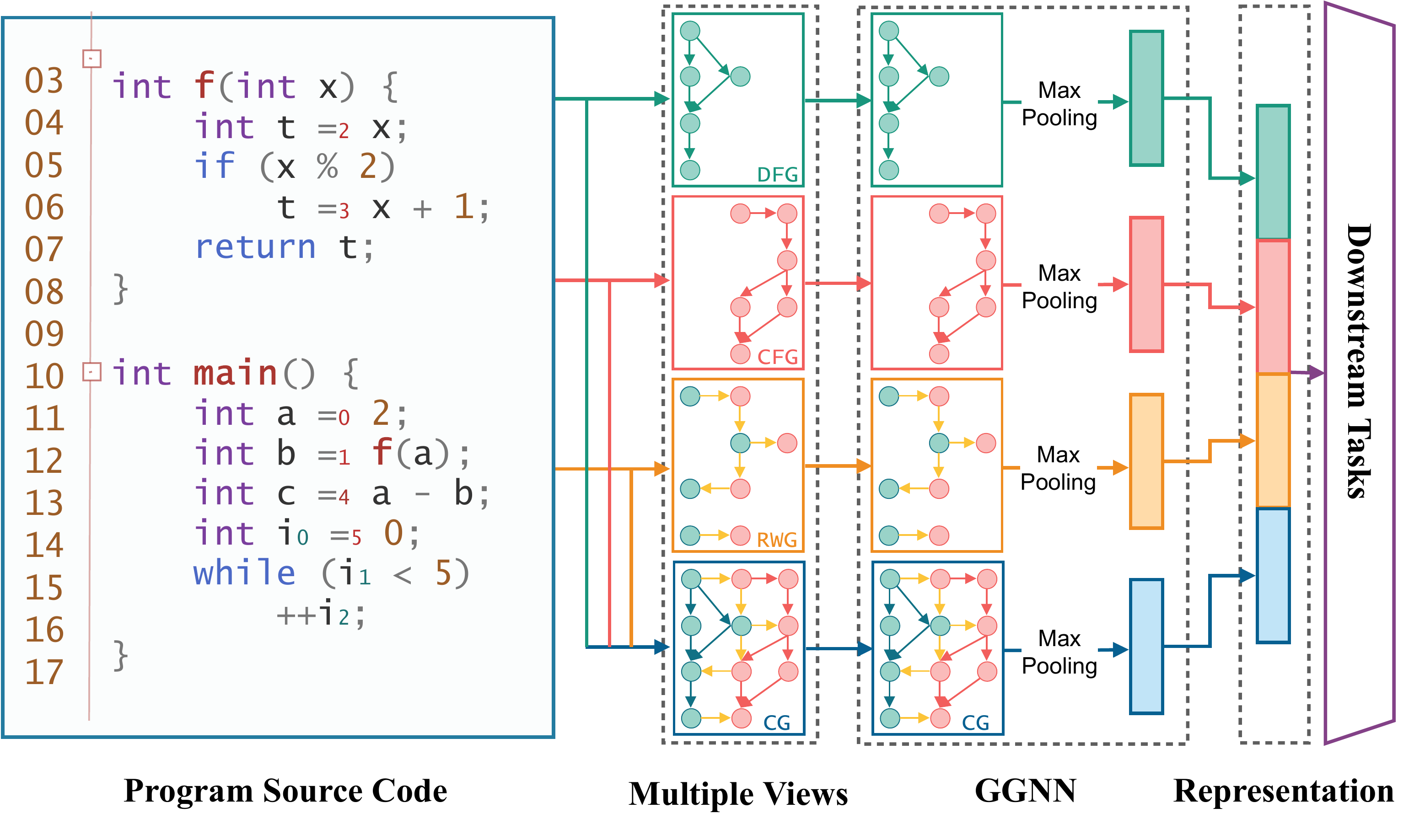}
    }
    \hspace{0.03\textwidth}
    \subfigure[An example of the combined graph.]{
        \centering
        \hspace{-10pt}
        \includegraphics[width=0.35\textwidth]{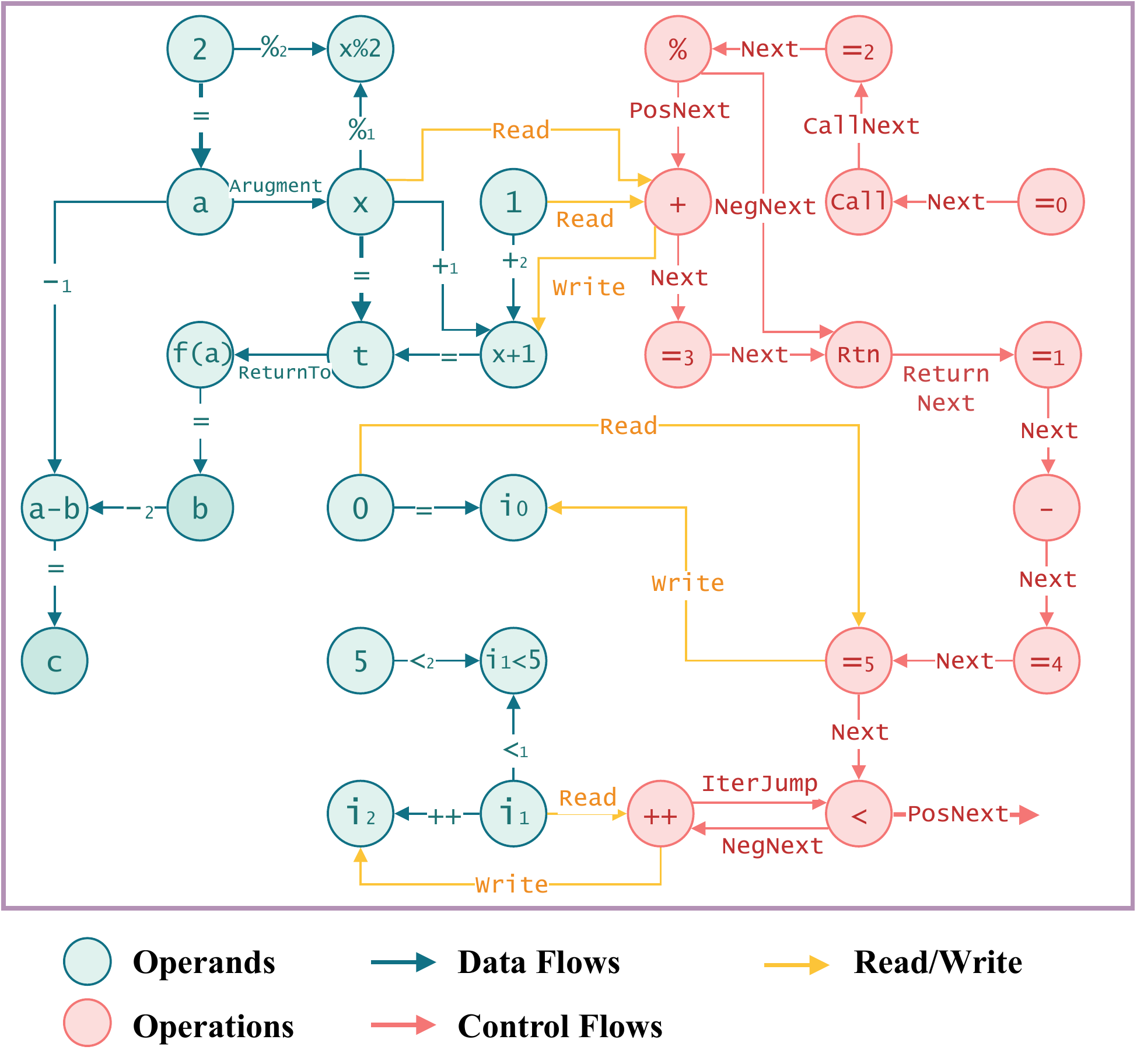}
    }
    \centering
    \caption{(a) The pipeline of MVG. Four graphs (\ie, DFG, CFG, RWG, and CG) are constructed based on the given source code. These constructed graphs are then fed into a GGNN to obtain a final program presentation for downstream tasks. 
    (b) An example of the combined graph (CG) corresponding to the program source code in (a).
    }
    \label{fig:pipeline}
\end{figure*}

This section describes how the MVG method converts programs into representational vectors. In particular, as displayed in Figure \ref{fig:pipeline}, we first represent a piece of source code as graphs of multiple views. We process these graphs with a gated graph neural network (GGNN) to extract the information in graphs. The extracted information is then combined to obtain a comprehensive representation. 

\subsection{Program Graphs of Multiple Views}
\label{sec:views}

To understand a program from different aspects and levels, we represent the program as graphs of multiple views. 
We consider four views in total:
\begin{inparaenum}[(1)]
    \item data-flow graph (DFG);
    \item control-flow graph (CFG);
    \item read-write graph (RWG); and
    \item combined graph (CG).
\end{inparaenum}

\subsubsection{Data-flow graph (DFG)}

Data flow is widely used to depicts programs in traditional program analysis \cite{aho1986compilers,farrow1976graph,fraser1995retargetable,muchnick1997advanced}. 
We use DFG to capture the relationship between operands. In DFG, nodes are
operands and edges indicate data flows.
As it is presented in Figure \ref{fig:pipeline}(b), the DFG of the code in Figure \ref{fig:pipeline}(a) is the green.
DFG includes two types of nodes, namely \emph{non-temporary operands}  and \emph{temporary operands}. Non-temporary operands denote variables and constants that explicitly exist in the source code, and temporary operands stand for temporary variables that only exist in program execution. 
Two groups of edges are considered:

\begin{itemize}
    \item \textit{Operation edges} exist in non-function-switch code. They connect the nodes to be operated and the nodes that receive the operation results.
    Standard operations are included in this category, 
    \eg, \texttt{=}, \texttt{+}, \texttt{-}, \texttt{*}, \texttt{/}, \texttt{>}, \texttt{<}, \texttt{==}.
    We distinguish different types of operations by using various types of edges. 
    \item \textit{Function edges} indicate data flows for function calls and returns, including two types of edges: \texttt{Argument} and \texttt{ReturnTo}. We use \texttt{Argument} edges in function calls to connect actual arguments and the corresponding formal arguments. We use \texttt{ReturnTo} edges to associate return values and the variables that receive the returns.
\end{itemize}

\subsubsection{Control-flow graph (CFG)}

We utilize CFG to model the execution order of operations. 
As Figure \ref{fig:pipeline}(b) shows, the CFG of the code in Figure \ref{fig:pipeline}(a) is the red. 
Based on compilers principles \cite{aho1986compilers,allen1970control}, we slightly adjust the design of CFG to better capture the key information of the program. 
Nodes in CFG are operations in the source code, including \emph{standard operations}, \emph{function calls} and \emph{returns}. 
Edges indicate the execution order of operations. The following edge types are considered: 

\begin{itemize}
    \item \textit{Condition edges} indicate conditional jumps in loops or branches (\eg, \texttt{while}, \texttt{for}, \texttt{if}). We define \texttt{PosNext} and \texttt{NegNext} two subtypes to represent situations where the conditions are \texttt{True} or \texttt{False} respectively. These edges start from condition operations and end at the first operation in the \texttt{True} or \texttt{False} blocks. 
    \item \textit{Iteration edges} are denoted as \texttt{IterJump}. We use them in loops (\eg, \texttt{while} and \texttt{for}) to indicate jumps at the end of each iteration, connecting the last and the first operations in the loop.
    \item \textit{Function edges} are used in function calls and returns, including two subtypes \texttt{CallNext} and \texttt{ReturnNext}. \texttt{CallNext} edges start from function call operations and point to the first operations in the called functions. \texttt{ReturnNext} edges begin with the last operations in called functions and end at the operations right after the corresponding function calls. 
    \item \textit{Next edges} stand for the most common execution order except for the above cases. Denoted as \texttt{Next}, they connect operations and their successor in execution order.
\end{itemize}

\subsubsection{Read-write graph (RWG)}
We design the RWG to capture the interaction between operands and operations.
As Figure \ref{fig:pipeline}(b) shows, part of RWG for the code in Figure \ref{fig:pipeline}(a) is the yellow edges and the nodes connect to yellow edges.
RWG is a bipartite graph with \emph{operands} and \emph{operations} as nodes. 
Two types of edges are introduced to connect operands and operations:

\begin{itemize}
    \item \textit{Read edges} start from operands and point to operations, meaning operations take operands to compute.
    \item \textit{Write edges} start from operations and point to operands, meaning variables receive the operation results.
\end{itemize}

\subsubsection{Combined graph (CG)}

In addition to DFG, CFG, and RWG, we further introduce a combined graph to capture the comprehensive overall information of a program. 
CG is an integral representation of the above three graphs and is obtained by first including all nodes and edges in DFG and CFG, and then adding \texttt{Read} and \texttt{Write} edges to connect variable and operation nodes as Figure \ref{fig:pipeline}(b) shows. 

~\indent

To summarize, formally, we can denote the graph of each view as $\mathcal{G}_i = \{ \mathcal{V}_i, \mathcal{E}_i\}$ where $i \in \{\text{DFG, CFG, RWG, CG}\}$, $\mathcal{V}_i$ is the node set and $\mathcal{E}_i$ is the edge set. We have $\mathcal{V}_{\text{RWG}}\subseteq\mathcal{V}_{\text{CG}}=\mathcal{V}_{\text{DFG}}\cup\mathcal{V}_{\text{CFG}}$, 
and
$\mathcal{E}_{\text{CG}}=\mathcal{E}_{\text{DFG}}\cup\mathcal{E}_{\text{CFG}}\cup\mathcal{E}_{\text{RWG}}$.

\subsection{Extracting Information with a GGNN}

As mentioned above, a program can be represented as four views in the form of graphs. Here, we adopt a gated graph neural network (GGNN) \cite{li2016gated}, a widely used graph neural network (GNN) model, to extract features from each graph view. 

For a graph $\mathcal{G}_i$ of an arbitrary view, this GGNN first initializes nodes' hidden representations with one-hot encodings of node types (\ie, operation or operand types). That is, for any node $u \in \mathcal{V}_i$, we initialize its hidden state as below:

\begin{equation}
    \textbf{h}_u^{0} = \textbf{x}_u,
\end{equation}
where $\textbf{h}_u^{0}$ is the initial hidden state of $u$, and $\textbf{x}_u$ is the one-hot encoding of $u$'s node type. 

The nodes then update their states by propagating messages in the graph as the following equations: 

\begin{align}
    \label{eq:mt}
    \textbf{m}_{u,v}^t 
    &= f_e(\textbf{h}_v^{t-1}), \quad e=(u,v)\in\mathcal{E}_i, \\
    \bar{\textbf{m}}_u^t 
    &= \text{Mean}(\{\textbf{m}_{u,v}^t\}_{v \in \mathcal{N}(u)}), \quad u\in\mathcal{V}_i, \\
    \label{eq:ht}
    \textbf{h}_u^t 
    &= \text{GRU} (\textbf{h}_u^{t-1},\bar{\textbf{m}}_u^t,), \quad u\in\mathcal{V}_i,
\end{align}

where 
$u,v$ are node indicators and $e$ is the edge which connects $u$ and $v$, 
$\textbf{m}_{u,v}^t$ stands for the message $u$ receives from $v$ at the $t$-th iteration, 
$f_e(\cdot)$ is a message passing function that depends on the edge type of $e$,
$\textbf{h}_v^{t-1}$ represents the hidden state of $v$ from the last iteration, 
$\bar{\textbf{m}}_u^t$ is the aggregated message received by $u$, 
$\text{Mean}(\cdot)$ denotes the average pooling function, 
$\mathcal{N}(u)$ is the set of $u$’s neighbors, 
$\textbf{h}_u^t$ is the updated hidden state,
and $\text{GRU}(\cdot)$ is a gated recurrent unit \cite{cho2014properties}. 

After $T$ iterations, the hidden states will contain enough information of the given graph. Therefore, we take the hidden states of nodes at the final iteration and integrate them using a max pooling
to obtain a final vector representation of the graph view $\mathcal{G}_i$:

\begin{equation}
    \label{eq:zi}
    \textbf{z}_i = \text{MaxPooling}(\{\textbf{h}_u^T\}_{u \in \mathcal{V}_i}). 
\end{equation}

\subsection{Program Representation}

To form an overall representation of the program, we concatenate representations from all views:

\begin{equation} 
\label{eq:prep}
    \textbf{z} = \textbf{z}_\text{DFG} \oplus \textbf{z}_\text{CFG} \oplus \textbf{z}_\text{RWG} \oplus \textbf{z}_\text{CG},
\end{equation}

where $\textbf{z}_\text{DFG}, \textbf{z}_\text{CFG}, \textbf{z}_\text{RWG},  \textbf{z}_\text{CG}$ are representations for DFG, CFG, RWG, and CG respectively computed as Equation \ref{eq:zi}, $\oplus$ denotes concatenation.

\indent

In summary, our proposed MVG method is outlined in Algorithm \ref{algo:process}. 

\renewcommand\arraystretch{0.7}

\vspace{10pt}
\begin{algorithm}[t]
    \caption{MVG Program Representation Method}
    \textbf{Input:} Source code of a program; \\
    \textbf{Output:} The vector representation of the input program;
    \label{algo:process}
    \vspace{-10pt}
    \begin{algorithmic}[1]
        \STATE Construct DFG, CFG, and RWG;
        \STATE Construct CG based on DFG, CFG, and RWG;
        \FOR{$\mathcal{G}_i\in\{\mathcal{G}_{DFG}, \mathcal{G}_{CFG}, \mathcal{G}_{RWG}, \mathcal{G}_{CG}\}$} 
            \STATE For $\forall u \in \mathcal{V}_i$, initialize its hidden representation with the one-hot encoding of the node type: $\textbf{h}_u^0=\textbf{x}_u$;
            \STATE Iteratively update node hidden representations with a  GGNN for $T$ steps (Eq.~\ref{eq:mt}-\ref{eq:ht});
            \STATE Compute the graph representation $\textbf{z}_i$ as Eq.~\ref{eq:zi};
        \ENDFOR
        \STATE Compute the program representation \textbf{z} as Eq.~\ref{eq:prep};
        \STATE Feed \textbf{z} to downstream tasks, \eg, algorithm detection;
    \end{algorithmic}
\end{algorithm}

\section{Experiments}

In this section, we evaluate our proposed MVG method on two algorithm detection datasets \texttt{POJ-104} and \texttt{ALG-109}. 
The implementation for our proposed MVG model and the datasets are available at \url{https://github.com/githubg0/mvg}.

\subsection{Baselines}

We compare our MVG method with four representative program representation methods in the recent literature. 
\begin{itemize}
    \item \textbf{NCC} \cite{ben2018neural} is a sequence-based method that compiles programs into intermediate representations (IRs) and obtains program representations with the skip-gram algorithm.
    \item \textbf{TBCNN} \cite{mou2016convolutional} is a tree-based method that extracts  features from program ASTs.
    \item \textbf{LRPG}\footnote{LRPG: In the published paper, this model is called GGNN, which may be confused with gated graph neural networks. Here we refer to it as LRPG by taking the abbreviation of
    the paper title.} \cite{allamanis2018learning} is a graph-based method. It introduces semantic edges such as control flows and data dependencies into the AST and extracts program features from the resulted graph.
    \item \textbf{Devign} \cite{zhou2019devign} is an extension of LRPG and improves the performance by including more types of control-flow and data dependency edges.
\end{itemize}

\subsection{\texttt{POJ-104}:
Algorithmic Problem Classification}

\subsubsection{Dataset description}

\texttt{POJ-104} is a public dataset that contains source code solutions for algorithmic programming problems on the Peking University online judge\footnote{Peking University online judge (POJ): \url{http://poj.org/}.} \cite{mou2016convolutional}. 
This dataset contains 52,000 programs, and each program is labeled with an algorithmic problem ID. 
In total, 104 problems are included, corresponding to a multi-class single-label classification problem with 104 classes. 

Typically, a particular algorithmic problem will require the solution code to contain certain algorithms or data structures to obtain the correct answer. 
Therefore, there is an implicit mapping between the problem ID labels and algorithm types. 
The statistics for this dataset are listed in Table \ref{tab:data}. 

\subsubsection{Implementation details}

We implement a rule-based parser to pre-process the source code of the input programs to obtain DFG, CFG, RWG, and we merge DFG, CFG, and RWG to generate the CG. To predict the label of the input programs, we feed its program representation to a two-layer multilayer perceptron (MLP) wrapped by the Softmax function. The dimension is selected from $\{100, 120, 140, 160, 180, 200\}$, the iterations $T$ for message propagation is selected from $\{1, 2, 4, 8\}$. We use the Adam optimizer \cite{kingma2014adam} to train the model, the learning learning rate is selected from $\{1 \times 10^{-3}, 6 \times 10^{-4}, 3 \times 10^{-4}, 1 \times 10^{-4}\}$. 
For all the baselines, the
hyperparameters are carefully tuned to the best performance.

\newsavebox{\tablebox}
\renewcommand\arraystretch{1}
\begin{table}[t]
    \centering
    \caption{Dataset statistics.}
    \label{tab:data}
    \scriptsize
    \begin{lrbox}{\tablebox}
    \hspace{-10pt}
    \begin{tabular}{p{1.8cm}|r|r|r}
        \hline\hline
         & \texttt{POJ-104} & \texttt{ALG-109} & \texttt{ALG-10} \\
        \hline
        Classification & Single-label & Multi-label & Multi-label \\
        \hline
        Label & Problem ID & Algorithms & Algorithms \\
        \hline
        \#Classes & 104 & 109 & 10  \\
        \hline
        \#Samples & 52,000 & 11,913 & 7,974 \\
        \hline
        Average \#lines & 36.26 & 94.27 & 94.37 \\
        \hline
        Average \#labels & 1.00 & 1.94 & 1.70\\
        \hline
        Language & \texttt{C} & \texttt{C}/\texttt{C++} & \texttt{C}/\texttt{C++} \\
        \hline\hline
    \end{tabular}
    \end{lrbox}
    \scalebox{1.2}{\usebox{\tablebox}}
\end{table}

\renewcommand\arraystretch{1}
\begin{table}[t]
    \centering
    \caption{Experiment results on \texttt{POJ-104}.}
    \label{tab:poj}
    \scriptsize
    \begin{lrbox}{\tablebox}
    \hspace{-10pt}
\begin{tabular}{c|c|c|c|c|c}
        \hline\hline
        \textbf{Method} & NCC &TBCNN & LRPG & Devign & \textbf{MVG} \\
        \hline
        \textbf{Accuracy(\%)} & 94.83 & 94.00 & 90.31 & 92.82  & \textbf{94.96} \\
        \hline\hline
    \end{tabular}
    \end{lrbox}
    \scalebox{1.2}{\usebox{\tablebox}}
\end{table}

\subsubsection{Results and discussion}

Following previous work \cite{mou2016convolutional,bui2021treecaps}, we evaluate the accuracy of model predictions on \texttt{POJ-104}. 
The higher accuracy denotes better performance. The experiment results are shown in Tables \ref{tab:poj}. 

From the results, we can see that MVG achieves the highest accuracy 94.96\%. However, other baselines can also achieve very high accuracy, \eg, 94.83\%, and 94.00\%. 
We assume this is because algorithmic problem classification is too easy for the models. For example, algorithmic problems will often require certain input and output formats, and this could leak information to the models, providing them a shortcut to classify the problem ID. 
Therefore, we do need a more challenging dataset to further distinguish the program representation power of models.

\subsection{\texttt{ALG-109}:
Algorithm Classification} 

\subsubsection{Dataset description}

\renewcommand\arraystretch{1}
\begin{table}[t]
    \centering
    \caption{Most frequent ten algorithms in \texttt{ALG-109}, denoted as \texttt{ALG-10}.}
    \label{tab:alg-10}
    \scriptsize
    \begin{lrbox}{\tablebox}
    \hspace{-10pt}
    \begin{tabular}{r|c|c |r|c|c}
            \hline\hline
            & \textbf{Algorithm} & \textbf{\#Samples} & & \textbf{Algorithm} & \textbf{\#Samples} \\
            \hline
            1 & \texttt{Recursion} & 4365 & 6 & \texttt{Enumeration} & 681 \\ \hline
            2 & \texttt{DepthFirstSearch} & 3117 & 7 & \texttt{GreedyAlgorithm} & 557 \\ \hline
            3 & \texttt{BreadthFirstSearch} & 1407 & 8 & \texttt{Recurrence} & 551 \\ \hline
            4 & \texttt{Queue} & 1083  & 9 & \texttt{DisjointSetUnion} & 548 \\ \hline
            5 & \texttt{SegmentTree} & 775  &10 & \texttt{QuickSort} & 501 \\
            \hline\hline
           
        \end{tabular}
    \end{lrbox}
    \scalebox{0.9}{\usebox{\tablebox}}
\end{table}

As mentioned above, the algorithmic problem classification dataset \texttt{POJ-104} is too easy to distinguish the representation power of compared models. 
Besides, there is no other public annotated algorithm detection dataset in the literature. 
Therefore, we construct a more realistic and more challenging algorithm classification dataset \texttt{ALG-109} by ourselves to serve as a new benchmark. 
\texttt{ALG-109} contains 11,913 pieces of source code collected from the the CSDN website\footnote{The CSDN website: \url{https://www.csdn.net/}.}. Each program is labeled with the algorithms and data structures that appear in the source code. So different from \texttt{POJ-104}, the \texttt{ALG-109} dataset corresponds to a much harder multi-class multi-label classification problem. 
The algorithm labels are annotated by previous programming contest participants who have adequate domain knowledge. 
Overall, 109 algorithms and data structures are considered. The most frequently appearing ten algorithms are listed in Table \ref{tab:alg-10}, and we denote this subset as \texttt{ALG-10}. The statistics of the constructed dataset are listed in Table \ref{tab:data}. 
We randomly split 80\% data for training and validation, and 20\% for testing. 

\subsubsection{Implementation details}

We implement a rule-based parser to pre-process the code to obtain DFG, CFG, and RWG, and we merge DFG, CFG, and RWG to obtain the CG. To predict the algorithms in the programs, we feed program representation to a two-layer MLP wrapped by a Sigmoid function to obtain the occurrence probability of each algorithm. If the occurrence probability of an algorithm is larger than 0.5, we consider it as one of the algorithms which implement the corresponding program. The dimension is selected from $\{120, 144, 168, 192, 216\}$, the iterations $T$ for message propagation is selected from $\{1, 2, 4, 8\}$. We use the Adam optimizer \cite{kingma2014adam} to train the model, the learning learning rate is selected from $\{1 \times 10^{-3}, 6 \times 10^{-4}, 3 \times 10^{-4}, 1 \times 10^{-4}\}$. For the baselines, the
hyperparameters are carefully tuned to the best performance.

\subsubsection{Results and discussion} 

\renewcommand\arraystretch{1}
\begin{table}[t]
    \centering
    \caption{Experiment results on \texttt{ALG-109} and \texttt{ALG-10}.}
    \label{tab:alg}
    \scriptsize
    \begin{lrbox}{\tablebox}
    \begin{tabular}{c|c|c|c|c} 
        \hline \hline 
        & \textbf{Method} & \textbf{Micro-F1(\%)} & \textbf{Exact Match(\%)} & \textbf{Ham-Loss(\%)} \\
        \hline
        \multirow{4}{*}{\rotatebox{90}{\texttt{ALG-109}}} 
        & NCC & 48.96 $\pm$ 0.91 & 21.01 $\pm$ 1.24 & 1.61 $\pm$ 1.24 \\
        & TBCNN & 35.03 $\pm$ 3.54 & 9.13 $\pm$ 1.34 & 1.44 $\pm$ 0.01\\
        & LRPG & 60.56 $\pm$ 0.87 & 30.14 $\pm$ 1.33 & 1.09 $\pm$ 0.02 \\
        & Devign & 56.90 $\pm$ 1.57 & 27.67 $\pm$ 1.04 & 1.16 $\pm$ 0.02\\
        & \textbf{MVG} & \textbf{65.26 $\pm$ 0.85} & \textbf{36.27 $\pm$ 0.67} & \textbf{1.03 $\pm$ 0.02} \\
        \hline
        \multirow{4}{*}{\rotatebox{90}{\texttt{ALG-10}}} 
        & NCC & 72.18 $\pm$ 0.89 & 46.46 $\pm$ 1.34 & 9.29 $\pm$ 0.28 \\
        & TBCNN & 67.53 $\pm$ 0.79 & 34.34 $\pm$ 0.96 & 9.88 $\pm$ 0.46\\
        & LRPG & 78.48 $\pm$ 1.51 & 55.21 $\pm$ 2.85 & 7.31 $\pm$ 0.59 \\
        & Devign & 78.40 $\pm$ 0.98 & 55.85 $\pm$ 1.88 & 7.16 $\pm$ 0.23 \\
        & \textbf{MVG} & \textbf{80.15 $\pm$ 0.86} & \textbf{58.36 $\pm$ 1.99} & \textbf{ 6.67  $\pm$ 0.29 }\\
        \hline \hline
    \end{tabular}
    \end{lrbox}
    \scalebox{1.1}{\usebox{\tablebox}}
\end{table}

We evaluate the performance of models on the testing data with three different metrics: the micro-F1 score, the exact match accuracy, and the Hamming loss. 
A higher micro-F1 score and exact match accuracy indicate a superior performance, while a lower Hamming loss stands for the better. 
The experiment results on \texttt{ALG-109} and \texttt{ALG-10} are shown in Table \ref{tab:alg}. 

From Table \ref{tab:alg}, we observe that:
\begin{inparaenum}[(1)]
    \item Our proposed MVG method surpasses all the baselines significantly on both \texttt{ALG-109} and \texttt{ALG-10}, illustrating MVG's superior performance on algorithm classification.
    \item Graph-based methods (\ie, LRPG, Devign, and MVG) all perform remarkably better than the sequence-based method (\ie, NCC) and the tree-based method (\ie, TBCNN), showing the great potential of representing programs as graphs.
    \item All models' performances drop when moving from \texttt{ALG-10} to \texttt{ALG-109}, because labels in \texttt{ALG-10} will be bound more training data. However, we can see the gap of MVG is smaller than others, which means MVG is relatively less sensitive to the insufficiency of data. 
    \item Comparing with the experiment results from \texttt{POJ-104}, we find the methods are more distinguishable on \texttt{ALG-109}, and there is still large room for models to further improve their performances on this dataset.    Therefore, our constructed \texttt{ALG-109} dataset might serve better as an algorithm detection or PLP program representation benchmark. 
\end{inparaenum}

\begin{figure}[t]
    \centering
    \hspace{-18pt}
    \includegraphics[width=0.5\textwidth]{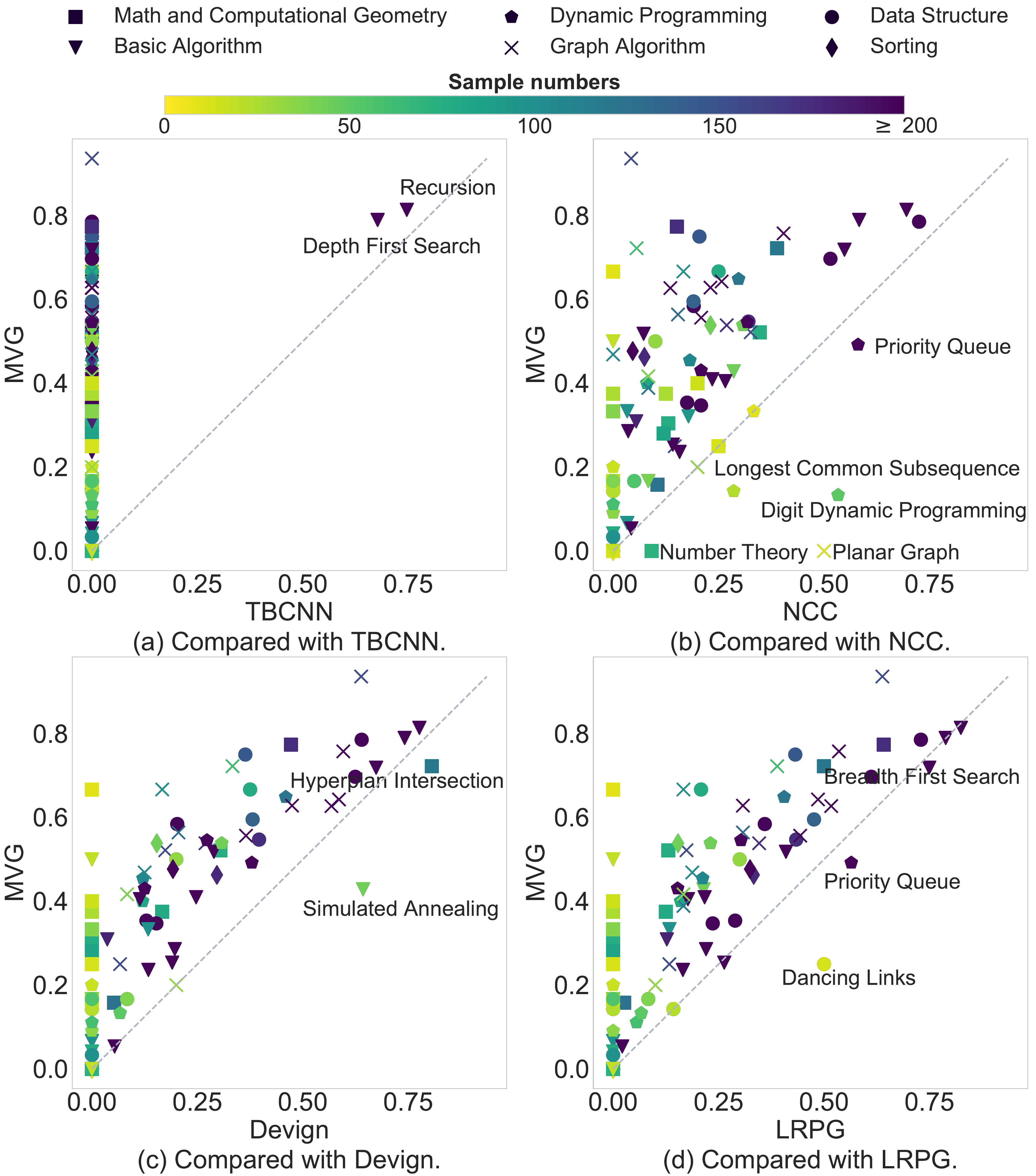}
    \caption{Comparing MVG and baselines on algorithm labels. In each subplot, each point represents one particular algorithm label. The $y$-axes are accuracy obtained by MVG, while the $x$-axes are accuracy obtained by the baselines. The number of samples and the algorithm type of each label are distinguished by the color and point style respectively. A point lying above $y=x$ means MVG is performing better than the baseline for this algorithm label. 
    }
    \label{fig:scatter}
\end{figure}

To further investigate how these models perform dissimilarly on each specific algorithm label, we compare MVG with the baselines and visualize the results in Figure \ref{fig:scatter}. 
From the visualization, we can see that, for almost all algorithm labels, MVG will perform superior to the baselines, especially for the labels with insufficient data. 

\subsubsection{Ablation study}

\renewcommand\arraystretch{1}
\begin{table}[t]
    \centering
    \caption{Ablation study on \texttt{ALG-109}.}
    \vspace{-10pt}
    \label{tab:abl}
    \scriptsize
    \begin{lrbox}{\tablebox}
    \begin{tabular}{l|c|c|c}
        \hline\hline
        \textbf{Variant} & \textbf{Micro-F1(\%)} & \textbf{Exact Match(\%)} & \textbf{Ham-Loss(\%)} \\
        \hline
        \textbf{MVG} & \textbf{65.26 $\pm$ 0.85} & \textbf{36.27 $\pm$ 0.67} & \textbf{1.03 $\pm$ 0.02} \\
        -- DFG & 62.34 $\pm$ 1.11 & 32.67 $\pm$ 0.85 & 1.09 $\pm$ 0.02 \\
        -- CFG & 64.18 $\pm$ 0.86 & 34.72 $\pm$ 1.08 & 1.06 $\pm$ 0.02 \\
        -- RWG & 64.01 $\pm$ 1.06 & 35.00 $\pm$ 0.91 & 1.06 $\pm$ 0.03 \\
        -- CG & 64.38 $\pm$ 0.78 & 34.86 $\pm$ 0.93 & 1.06 $\pm$ 0.02 \\
        OnlyCG & 62.02 $\pm$ 0.74 & 32.06 $\pm$ 0.99 & 1.09 $\pm$ 0.02 \\
        $+$AST & 65.19 $\pm$ 0.94 & 36.10 $\pm$ 1.25 & 1.04 $\pm$ 0.02 \\
        \hline\hline
    \end{tabular}
    \end{lrbox}
    \scalebox{1.2}{\usebox{\tablebox}}
\end{table}

\begin{figure}[t]
    \centering
    \hspace{-15pt}
     \includegraphics[width=0.5\textwidth]{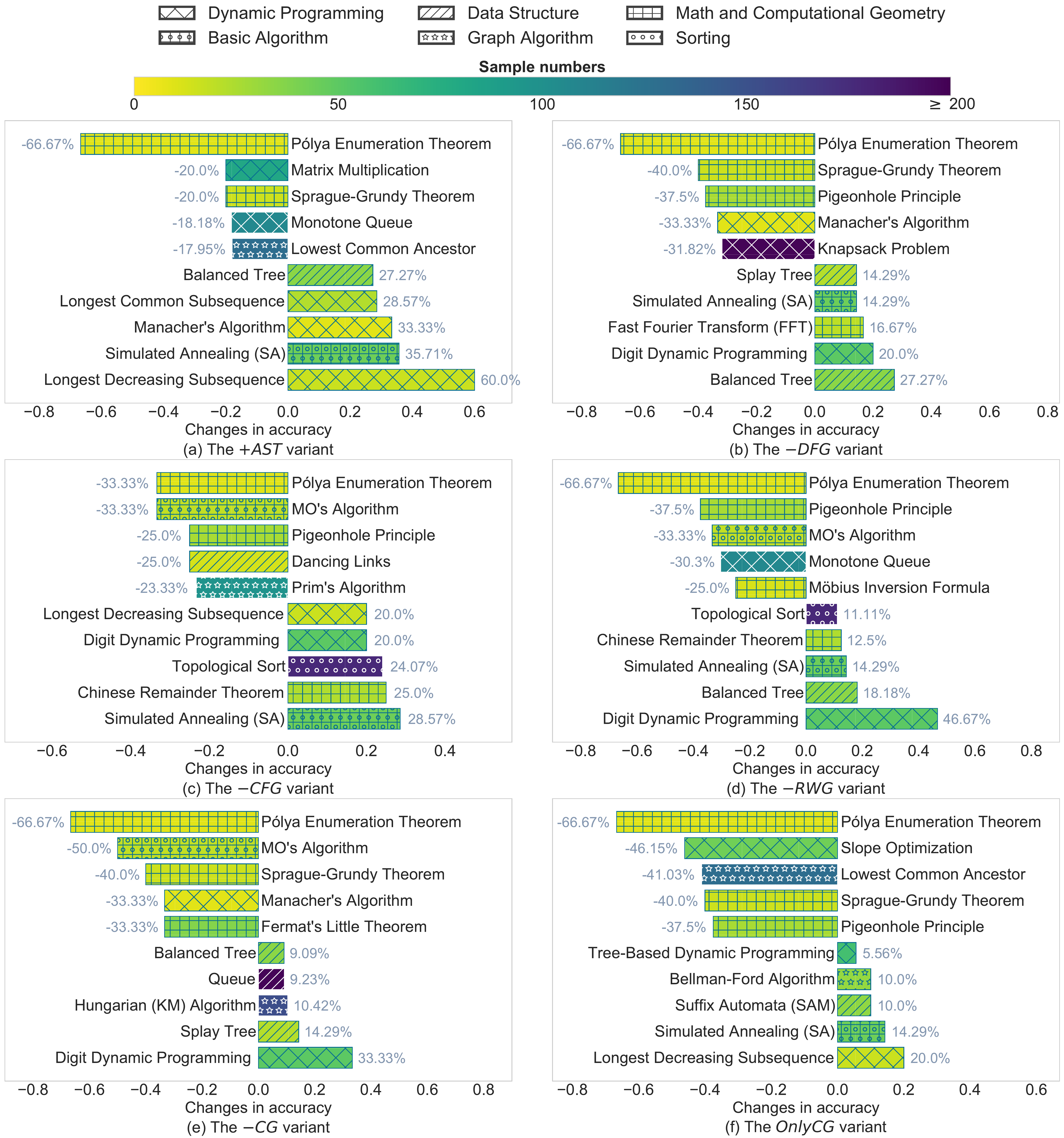}
    \caption{Algorithm classification accuracy changes of model variants. For each variant, the top fives labels with the largest performance drops as well as increases are shown. 
    }
    \label{fig:accuracy_constrast}
\end{figure}

To obtain a deep understanding of MVG's outstanding performance, we conduct some further ablation studies to learn each view's impact on the MVG model. Here, we consider six variants:

\begin{itemize}
    \item \textbf{-DFG} removes the DFG view from MVG. The data-flow information in CG is also removed accordingly.
    \item \textbf{-CFG} removes the CFG view from MVG. The control-flow information in CG is also removed accordingly.
    \item \textbf{-RWG} removes the RWG view from MVG. The read-write information in CG is also removed accordingly.
    \item \textbf{-CG} removes the combined CG view from MVG, so the other three views (\ie, DFG, CFG, RWG) will no longer interact with each other. 
    \item \textbf{OnlyCG} contains only the combined CG view, so the data-flow, control-flow, and read-write information will be mixed up together into one single view;
    \item \textbf{+AST} adds an abstract syntax tree (AST) view to MVG, which will include more syntactic information.
\end{itemize}

The ablation study results are listed in Table \ref{tab:abl}.
From the results, we find that:
\begin{inparaenum}[(1)]
    \item Removing any view from MVG (\ie, -DFG, -CFG, -RWG, and -CG) will cause a drop in performance, showing the indispensable role of every view in program representation.
    \item Adding the AST view (\ie, +AST) harms the performance slightly, which means the AST view is unnecessary in algorithm detection task and we should not emphasize too much on the syntax in program representation. 
    \item Removing CG undermines the accuracy, meaning interactively combining the other three views helps MVG to better understand the programs. 
    On the other hand, the performance of the OnlyCG variant is also inferior to MVG. Therefore, we can conclude that both the independent views (\ie, DFG, CFG, RWG) and the integral view (\ie, CG) are necessary for our program representation. 
    \item Comparing all the variants, we find by deleting the DFG view, the performance drops the most, showing that DFG is most critical in our program representation model.
\end{inparaenum}

We also examine how the performance of different algorithms changes when using different model variants. The results are displayed in Figure \ref{fig:accuracy_constrast}. Here, for each model variant, we show the top five labels with the largest performance drops and increases.
From the results, we observe that:
\begin{inparaenum}[(1)]
    \item Overall, by changing the design of MVG into other variants, the performance drops more while increasing less;
    \item Compared with the variants, our MVG seems to have better representation for programs that contain \textit{Math and Computational Geometry} algorithms, \eg., \texttt{Pólya Enumeration Theorem}, since when replacing MVG with other variants, the detection performance on these algorithms drop significantly.
\end{inparaenum}

\subsubsection{Case study}

\begin{figure}[t]
    \centering
    \vspace{-5pt}
    \subfigure[Sorting algorithms.]{
    \centering 
    \hspace{-10pt}
    \includegraphics[width=9cm]{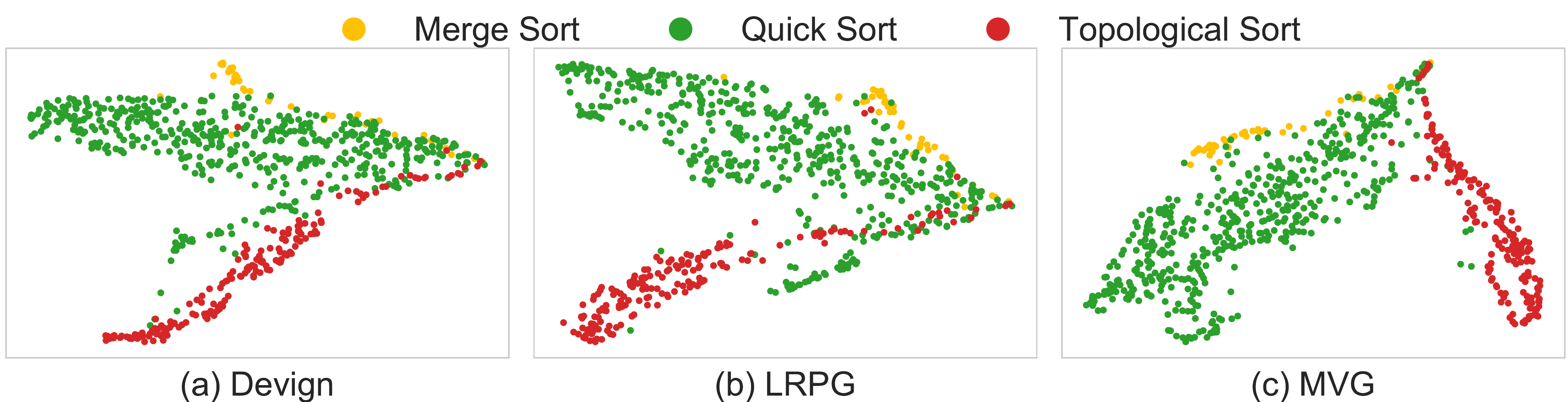}
    }
    \vspace{-5pt}
    \subfigure[Shortest path algorithms]{
    \centering
    \hspace{-10pt}
    \includegraphics[width=9cm]{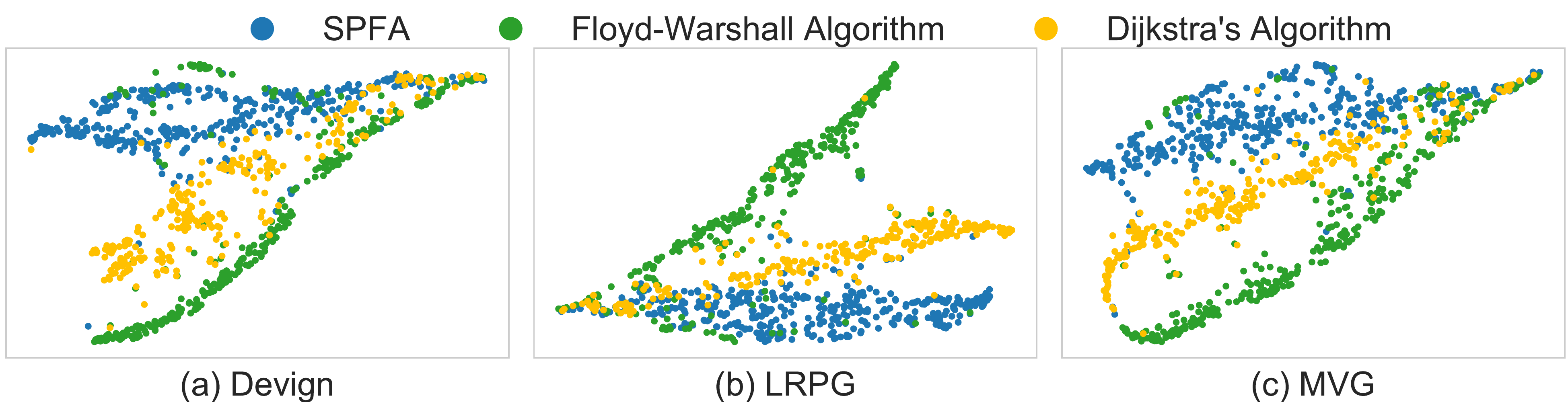}
    }
    \vspace{-5pt}
    \subfigure[\texttt{Dijkstra's Algorithm} and \texttt{Prim's Algorithm}. ]{
    \centering
    \hspace{-10pt}
    \includegraphics[width=9cm]{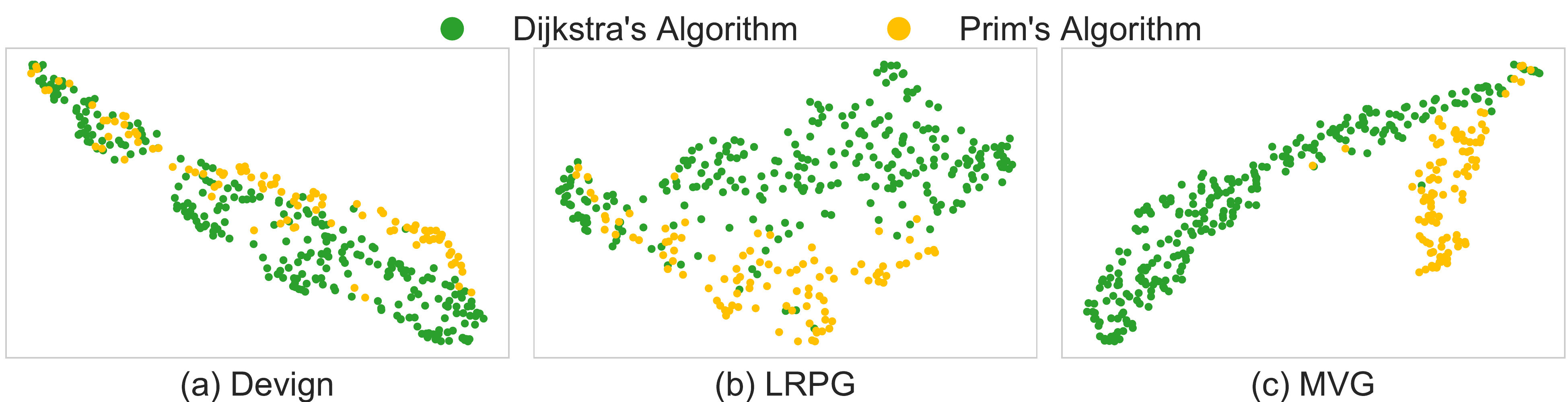}
    }
    \centering
    \caption{Visualization of program representations. } 
    \label{fig:contrast}
\end{figure}

To intuitively figure out whether our MVG model can render better program representations, we use UMAP \citep{mcinnes2018umap} to visualize the representation vectors encoded by the
three best-performing models (\ie, MVG, Devign, and LRPG) in Figure \ref{fig:contrast}. 

Three groups of algorithms are compared:
\begin{inparaenum}[(1)]
    \item We first compare three sorting algorithms (\ie, \texttt{Merge Sort}, \texttt{Quick Sort}, and \texttt{Topological Sort}). From the visualization, we can see that both Devign and LRPG can not distinguish \texttt{Quick Sort} and \texttt{Topological Sort} well, while MVG represents these two algorithms more differently. 
    \item For the shortest-path algorithms, the representation power of Devign, LRPG, and MVG seem almost the same. 
    \item We also compare  \texttt{Dijkstra's Algorithm} and  \texttt{Prim's Algorithm}, since they are designed for different intentions while having very similar implementations (\ie, both the two algorithms utilize the \texttt{Breadth-First Search}). From the visualization, we can see MVG gives a much more clear decision boundary of the two algorithms, meaning our method has higher representation power than the other two baselines. 
    \item Associating the results presented in Table \ref{tab:alg} and Figure \ref{fig:scatter}, we find MVG is more capable of representing source code especially under the context of algorithm detection. 
\end{inparaenum}

\section{Conclusion}

This paper presents a multi-view graph (MVG) program representation method for PLP. 
To understand source code more comprehensively and semantically, we propose to include four graph views of different levels and various aspects:
the data-flow graph (DFG), 
the control-flow graph (CFG),
the read-write graph (RWG), and
an integral combined graph (CG).
We evaluate our proposed MVG method in the context of algorithm detection, which is an important and challenging subfield of PLP. 
To fill the vacancy of a high-quality algorithm detection dataset, we construct \texttt{ALG-109}, an algorithm classification dataset that contains 109 algorithms and data structures in total. 
In experiments, MVG achieves state-of-the-art performance, demonstrating its outstanding capability of representing programs. 

For future work, it would be interesting to investigate how our MVG approach can be combined with other orthogonal techniques like pre-training. Moreover, we might also apply the MVG model and the annotated dataset \texttt{ALG-109} for the purpose of programming education.

\section{Acknowledgement}
We would like to thank Enze Sun and Hanye Zhao from Shanghai Jiao Tong University for their efforts in reviewing the data and baselines.
We also thank anonymous reviewers for their constructive comments and suggestions.
This work is partially supported by the Shanghai Municipal Science and Technology Major Project (2021SHZDZX0102) and the National Natural Science Foundation of China (62177033). 

\newpage
\vspace{-10pt}
\bibliography{mybib}

\end{document}